\def\year{2020}\relax
\documentclass[letterpaper]{article} 
\usepackage{aaai20}  
\usepackage{times}  
\usepackage{helvet} 
\usepackage{courier}  
\usepackage[hyphens]{url}  
\usepackage{graphicx} 

\usepackage{amsfonts,amssymb} 
\usepackage{algorithm,algcompatible,amsmath}

\algnewcommand\INPUT{\item[\textbf{Input:}]}%
\algnewcommand\OUTPUT{\item[\textbf{Output:}]}%
\usepackage{amsmath}

\usepackage{booktabs,ragged2e}

\usepackage[flushleft]{threeparttable}
\usepackage[utf8]{inputenc}
\usepackage{diagbox} 
\usepackage{longtable}
\usepackage{array}
\usepackage{rotating}
\usepackage{eqparbox}
\usepackage{makecell}
\usepackage{booktabs}
\usepackage{cleveref}

\usepackage[nocomma]{optidef}

\usepackage{graphicx}  
\title{Tensor Completion for Weakly-dependent Data on Graph \\ for Metro Passenger Flow Prediction}
\author{Ziyue Li\textsuperscript{\rm 1}\thanks{We show our great appreciation to the Metro Corporation for sharing this passenger flow data, and in the protection of privacy, all data has been desensitized.}, Nurettin Dorukhan Sergin\textsuperscript{\rm 2}\\ \Large \textbf{Hao Yan\textsuperscript{\rm 3}, Chen Zhang\textsuperscript{\rm 4}, Fugee Tsung\textsuperscript{\rm 5}}\\ 
\textsuperscript{\rm 1}\textsuperscript{,\rm 5}The Hong Kong University of Science and Technology\\ 
\textsuperscript{\rm 2}\textsuperscript{,\rm 3}Arizona State University,
\textsuperscript{\rm 4} Tsinghua University\\
}
 
\begin{document}

\maketitle

\begin{abstract}
Low-rank tensor decomposition and completion have attracted significant interest from academia given the ubiquity of tensor data. However, low-rank structure is a global property, which will not be fulfilled when the data presents complex and weak dependencies given specific graph structures. One particular application that motivates this study is the spatiotemporal data analysis. As shown in the preliminary study, weakly dependencies can worsen the low-rank tensor completion performance. In this paper, we propose a novel low-rank CANDECOMP / PARAFAC (CP) tensor decomposition and completion framework by introducing the $L_{1}$-norm penalty and Graph Laplacian penalty to model the weakly dependency on graph. We further propose an efficient optimization algorithm based on the Block Coordinate Descent for efficient estimation. A case study based on the metro passenger flow data in Hong Kong is conducted to demonstrate an improved performance over the regular tensor completion methods.
\end{abstract}

\section{Introduction}
\noindent Higher-order tensors have been actively used in research since they have an inclination to successfully preserve the complicated innate structural properties of data\cite{kolda2009tensor}. A tensor can be defined mathematically as the multi-dimensional arrays. The order of a tensor is the number of dimensions, also known as modes. 
Tensor completion, which is a missing data imputation task based on the observed data, has attracted significant amounts of research in areas such as image processing and machine learning. Specifically, spatiotemporal data can normally be modeled as high-order tensors with spatial and temporal modes; tensor analysis for spatiotemporal data  is widely used for prediction and feature extractions  \cite{sun2016understanding,dunlavy2011temporal}. Tensor completion can be also used as a prediction method, with the prediction horizon regarded as missing entries, and historical data as observed entries (Tan et al, 2016, Luan, Zhang 2018).

However, tensor completion of spatiotemporal data is a challenging task due to its complex spatial and temporal dependencies. Take the metro passenger flow tensor as an example, for temporal dependence, the prediction of a certain time is usually correlated with the historical observations. Passenger flow patterns also display a strong periodicity with period as 7 days.

For spatial dependency, graph structures are commonly used. For example, a geographical graph is often defined based on its spatially adjacent neighbours. Furthermore, a contextual graph is often defined based on the contextual similarity, which quantifies whether two stations share a similar function (e.g., business center, residential area or school etc.). In the relevant literature, it has been highlighted that both the geographical graph and contextual graphs significantly affect  passenger flow patterns at station points \cite{zhong2017spatiotemporal}. 

\begin{figure}[t]
\centering
\includegraphics[width=0.95\columnwidth]{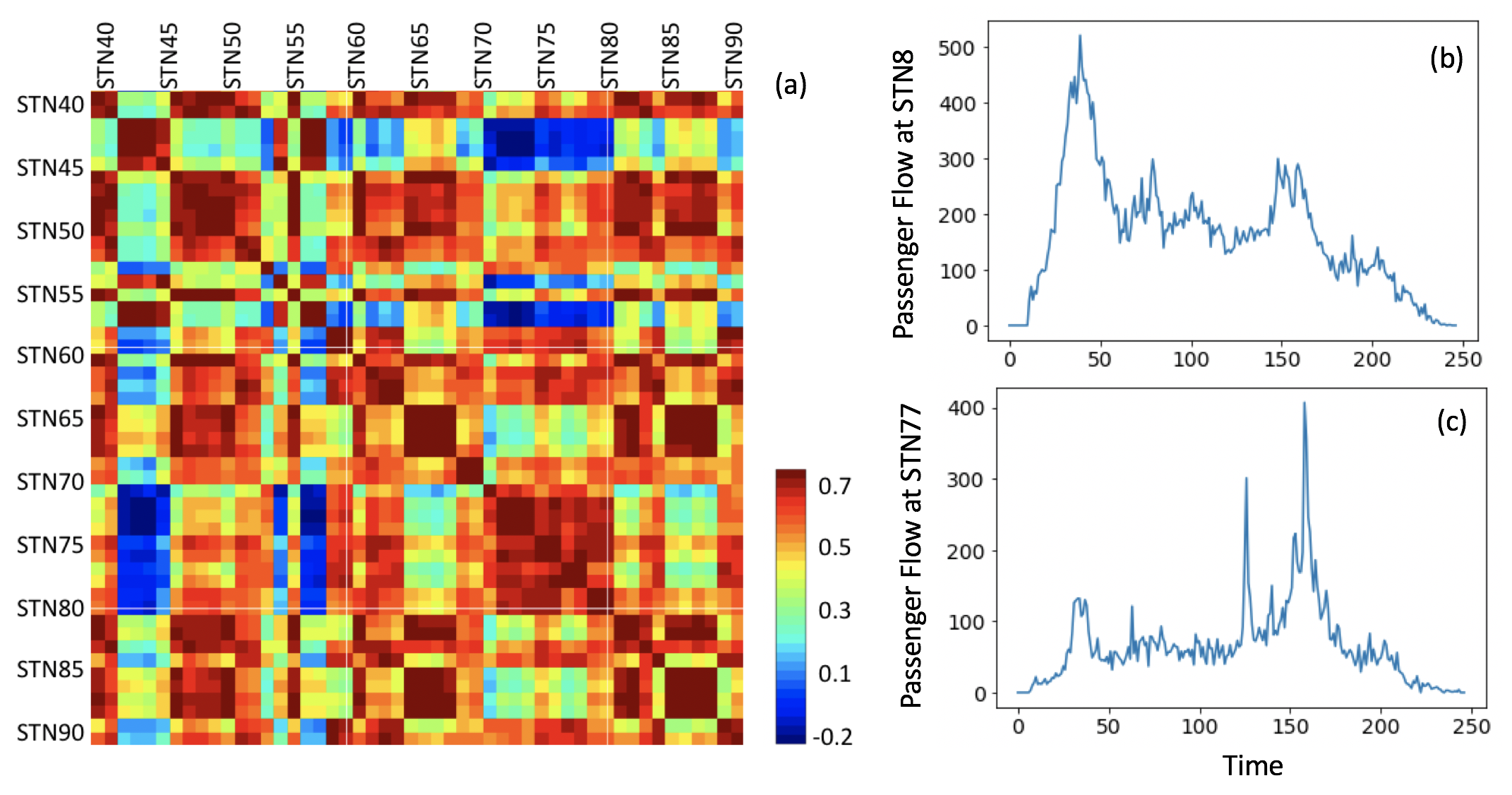} 
\caption{(a) The weakly-dependent property of each stations (b) Passenger Flow at Station 8 (c) Passenger Flow at Station 77}
\label{fig1}
\end{figure}

Taking the metro passenger flow data as example, denoted as three-mode tensor $\boldsymbol{\mathcal{X}}^{station \times  point \times day}$, as a result of the geographical and contextual graphs along the station mode, if any two stations are picked, they are less likely to be highly-dependent to each other if they are not connected in either of the two graphs, such as Station 8 and Station 77 in Figure 1(b), (c) having quite different passenger flow profile. With the amount of stations increasing, the chance that all stations are strongly dependent is very small. This data structure is defined as the weakly dependency on graphs, which has also been explored in other settings and applications \cite{zhang2018weakly}. 

Here, we like to give a formal definition of the \textit{Weakly-Dependent data on Graph (WDG)}. Intially, a graph or multiple graphs can be defined on the entities along a specific mode of a tensor. The tensor data is \textit{WDG} only if it is linear dependent on the neighborhood of the graph. We can further define this mode as a \textit{weakly-dependent mode}, and this data as \textit{weakly-dependent data}. For example, in the previous metro passenger flow example, the station is the weakly dependent mode. 

In the literature, the Low Rank Tensor Completion (LRTC) method is commonly used for a spatio-temporal data prediction, which fills in the incomplete tensor based on the low-rank tensor constraint. However, the low-rank property implies that data in all stations are linearly dependent, whereas the WDG is a local structure, which implies the data are only dependent on the neighborhood of the graph. Therefore, applying LRTC often yields an oversimplified model \cite{zhao2015bayesian}. 

In conclusion, we proposed a novel regularized CANDECOMP / PARAFAC (CP) tensor decomposition and completion framework considering the WDG property. It has been shown that the suggested framework is able to simultaneously achieve not only high accuracy considering both multi-networks but also a high degree of weak dependency. In summary, this paper makes following contributions:

\begin{itemize}
\item Formulate a Low Rank Tensor Completion with weakly-dependent pattern on graphs based on CP Decomposition.
\item Propose an efficient optimization algorithm for the proposed framework
\item Demonstrate the performance of the proposed framework in metro passenger flow data
\end{itemize}

\section{Literature Review}
\textbf{Tensor Decomposition and Regularization: }
Tensor decomposition has been widely studied in the available literature. Specifically, CANDECOMP / PARAFAC (CP) decomposition represents a tensor as the weighted summation of a set of rank-one tensors. For more detailed introduction to tensor decomposition and CP decomposition, please refer to  \citeauthor{kolda2009tensor}'s work (2009). 
\begin{equation}
\begin{split}
\mathcal{X} &=\sum_{r=1}^{R}\lambda_{r}  \mathbf{u}_{r}^{(1)}\circ \mathbf{u}_{r}^{(2)}\circ\cdots\circ \mathbf{u}_{r}^{(K)}\\
&=[[\lambda;\mathbf{U}^{(1)},\mathbf{U}^{(2)},...,\mathbf{U}^{(K)}]]
\end{split}
\end{equation}where $\mathbf{U}^{(k)}$ is the decomposed factor along $k$-th mode, which can be termed as the $k$-th mode factor.

To incorporate the domain knowledge into the tensor decomposition framework, constrained or regularized decomposition is commonly used. Some of the proposed frameworks include non-negative  tensor decomposition \cite{morup2008algorithms}, sparse tensor decomposition with  $L_{2,1}$-norm  penalty (Yu et al. 2019), and smooth tensor decomposition with total variation regularization  \cite{yokota2016smooth}, and a graph Laplacian penalty for tensor completion\cite{wang2015citywide}.


\textbf{Tensor completion: }
In terms of missing data imputation in a tensor, the state-of-the-art tensor completion methodology can be summarized into two major types. 

The first type is to define the low-rank penalty directly onto the original tensor. The benefit is that these frameworks are typically convex and can be optimized globally. One of the most popular methods involves the use of the tensor-version of the nuclear norm, which is defined as the sum of the nuclear norms of the all the unfolded matrices of $\mathcal{X}$ \cite{liu2012tensor} 
For example, Simple Low-Rank Tensor Completion (SiLRTC) defined the tensor completion problem as follows:

\begin{align}
\mathop{\mathrm{min}}_{\mathcal{X}, \mathbf{Y}_{(1)},...,\mathbf{Y}_{(K)}} & \sum_{k=1}^{K} \alpha_{k}\|\mathbf{Y}_{(k)}\|_{*}+\frac{\beta}{2}\| \mathbf{X}_{(k)}-\mathbf{Y}_{(k)} \|_{F}^{2} \\
 \textit{s.t. } & P_{\Omega}(\mathcal{X})=P_{\Omega}(\mathcal{T}) \nonumber
\end{align}

where $\mathcal{X}$ is the completed output tensor, $\mathcal{T}$ is the incomplete input tensor, and $\{\mathbf{Y}_{(k)}\}_{k=1,...,K}$ are the low-rank matrices corresponding to $k$-th mode unfolded tensor $\mathbf{X}_{(k)}$.

However, nuclear norm minimization is often very computationally expensive and typically involves singular value decomposition of very large matrices \cite{kressner2014low}.

Furthermore, another drawback of first type is that by using tensor decomposition, the decomposed factors usually carry some quite meaningful physical information which reflects a number of features in reality \cite{sun2016understanding}. Sun noted the peak hour pattern, dominant passenger type, highly-traveled area based on the decomposed factors along time, passenger type and location mode respectively. Inversely, if the auxiliary information from the physical world is available, then formulating a completion without a decomposition structure equated to being incapable of adding the prior information to the model. This shortage can be overcome by adopting a second approach, the completion based on a decomposition structure. With this, the prior information can be easily applied to the decomposed latent factors through proper regularization.

\textbf{Simultaneous Tensor Decomposition and Completion:}
The second approach is Simultaneous Tensor Decomposition and Completion (STDC), which is defined by \citeauthor{chen2013simultaneous}. STDC estimates the decomposed latent factors using partially observed data \cite{chen2019bayesian}. 
It is then possible to compute the missing entry from the estimated latent factors. 

In the STDC framework, factor prior or regularization is often applied on decomposed components based on low-rank structure \cite{chen2013simultaneous}. For example, in the available literature, a Bayesian hierarchical probabilistic structure for CP decomposition \cite{zhao2015bayesian} and Tucker decomposition (Zhao, Zhang and Cichocki 2016) was proposed, which assumed that the decomposed factor matrices were generated from a same high-level Gaussian distribution. Through this the same sparsity pattern can be obtained in latent matrices, yielding the minimum number of rank-one tensor . 
Low rank penalties such as $L_{1}$-norm penalty on the CP decomposition weight vector or nuclear-norm on the Tucker decomposition core tensor have also been proposed \cite{shi2017tensor}. 

However, as mentioned before, low-rank property can not be satisfied. Introducing the WDG into the tensor completion is  new challenge in the literature. 
In our preliminary result, we have shown that the Bayesian Low-Rank Tensor Completion performs much worse in weakly-dependent data than in strongly-dependent data. The WDG data thus presents a more diverse and complicated spatiotemporal dependency structure.

\section{Formulation}
We will formulate the tensor completion problem given the WDG data structure.

\subsection{Notations and Operations}

Through out this exposition, scalars are denoted in italics, e.g. $n$; vectors by lowercase letters in boldface, e.g. $\mathbf{u}$; and matrices by uppercase boldface letters, e.g. $\mathbf{U}$; High dimensional data, tensor by boldface script capital $ \boldsymbol{\mathcal{X}}$.

For a given incomplete tensor $ \boldsymbol{\mathcal{X}}\in \mathbb{R}^{{I}_{1}\times {I}_{2} \times ... \times {I}_{N}}$, we assume its first mode is WDG mode.

\subsection{CP Decomposition Based Tensor Completion}

\textbf{(1) Low-Rank Tensor Completion}

The low-rank tensor completion based on CP decomposition based on (1) can be formulated by $L_{1}$-norm on the weight vector:

\begin{align}
\begin{split} \mathop{\mathrm{min}}_{R, \boldsymbol{\mathcal{Y}}, \boldsymbol{\lambda}, {\{\mathbf{u}_{r}^{(k)}\}}} &
\frac{1}{2} \|  \boldsymbol{\mathcal{Y}}-\sum_{r=1}^{R}\lambda_{r}  \mathbf{u}_{r}^{(1)}\circ \mathbf{u}_{r}^{(2)}\circ\cdots\circ \mathbf{u}_{r}^{(K)}\|^{2} \\ 
& + \beta\|\boldsymbol{\lambda}\|_{1} \label{eq:cp}
\end{split} 
\\ \textit{s.t. } &  P_{\Omega}( \boldsymbol{\mathcal{Y}}) =P_{\Omega}( \boldsymbol{\mathcal{X}}) \nonumber
\\  & {\mathbf{u}_{r}^{(k)}}^{T} \mathbf{u}_{r}^{(k)} = 1,\quad \forall r \nonumber
\end{align}where $ \boldsymbol{\mathcal{X}}$ is the incomplete input tensor with the observation indicator tensor as $\Omega$, and $ \boldsymbol{\mathcal{Y}}$ is the complete output tensor. $\boldsymbol{\lambda}=[\lambda_{1}, \lambda_{2},...,\lambda_{R}]^{T}$  is the weight vector of CP decomposition. $L_{1}$-norm penalty on $\boldsymbol{\lambda}$ helps control the rank of the tensor. 

\noindent \textbf{(2) Weak Dependency}

\Cref{eq:cp} assumes a global low rank property within the entire tensor. The low-rank structure holds in general  when data are strongly-dependent. However, it does not hold for WDG data since this weak dependence or various features indicate that data may come from different sources or clusters. A relevant example here would be metro passenger flow pattern of different areas. Borrowing the idea of sparse coding \cite{lee2007efficient}, it is often more desirable to set $\mathbf{U}_{1}\in \mathbb{R}^{{I}_{1}\times R}$ to be sparse to force scores on unrelated entities to be zero \cite{zhang2018weakly}. 
Review the previous example of metro passenger flow tensor: based on their functionality (e.g., for commercial, entertainment, or residential use), all stations in the metro system can be regarded as coming from $c$ clusters, which have strong within-cluster dependence but little or no cross-cluster dependence. Directly using the low-rank tensor completion for all the $I_{1}$ stations will achieve poor completion and decomposition since the low-rank tensor completion may use features of stations from unrelated clusters and can eventually contaminate the tensor completion result. By introducing sparsity on $\mathbf{U}_{1}$, we represent a station's profile using data from the few selected stations (e.g., within the same cluster).  

\begin{align}
\mathop{\mathrm{min}}_{R, \boldsymbol{\mathcal{Y}}, \boldsymbol{\lambda}, {\{\mathbf{u}_{r}^{(k)}\}}} & loss  +\alpha\|\mathbf{U}_{1}\|_{1} + \beta\|\boldsymbol{\lambda}\|_{1} \\
\textit{s.t. } &  P_{\Omega}( \boldsymbol{\mathcal{Y}}) = P_{\Omega}( \boldsymbol{\mathcal{X}}) \nonumber\\
& {\mathbf{u}_{r}^{(k)}}^{T} \mathbf{u}_{r}^{(k)}=1, \quad \forall r \nonumber
\end{align}where $loss  =  \frac{1}{2} \| \boldsymbol{\mathcal{Y}}-\sum_{r=1}^{R}\lambda_{r}  \mathbf{u}_{r}^{(1)}\circ \mathbf{u}_{r}^{(2)}\circ\cdots\circ \mathbf{u}_{r}^{(K)}\|^{2}$. Besides, if the WDG data has multiple weakly-dependent modes, this $L_{1}$ penalty can also be easily extended into $\alpha_{1}\|\mathbf{U}_{1}\|_{1} +\alpha_{2}\|\mathbf{U}_{2}\|_{1}+\dots$

\noindent \textbf{(3) Graph Structure}

When auxiliary information is available, for example, the graph information of each entity in weakly-dependent mode is obtainable, then a more explicitly defined dependence across each entity can be formulated. For the graph $\mathbf{G}_{i}$, which contains $I_{1}$ entities or nodes inside, an corresponding Adjacency Matrix  $\mathbf{A}_{i} \in \mathbb{R}^{I_{1} \times I_{1}}$ can be obtained. According to the adjacency matrix, a penalty which force two entities close or similar to each other into sharing similar pattern is desired. With this in mind, a Laplacian penalty \cite{wang2015citywide,yu2019coupled} is a sensible choice. 
In Wang's work of citywide traffic tensor, the traffic pattern of different road segments depends on road segment correlation, so the penalty $tr(\mathbf{U}^{T}\mathbf{L}\mathbf{U})$ is added with $\mathbf{L}$ as the Laplacian Matrix of correlation matrix \cite{wang2015citywide}. 
Therefore, we define the Laplacian Matrix of $\mathbf{A}_{i}$ as $\mathbf{L}_{i}$. $\mathbf{L}_{i} = \mathbf{D}_{i} -\mathbf{A}_{i}$ and $\mathbf{D}_{i}$ is a diagonal matrix with element $d_{i_{1},i_{2}}=\sum_{i_{1}}a_{i_{1},i_{2}}$. Consequently, the formulation with graph information is:
\begin{align}
\begin{split}
\mathop{\mathrm{min}}_{R, \boldsymbol{\mathcal{Y}}, \boldsymbol{\lambda}, {\{\mathbf{u}_{r}^{(k)}\}}} & 
loss + \alpha\|\mathbf{U}_{1}\|_{1}  + \beta\|\boldsymbol{\lambda}\|_{1} + \frac{1}{2}\gamma\cdot \\ & + tr(\mathbf{U}_{1}^{T} \mathbf{L}_{i}\mathbf{U}_{1})
\end{split} \\ 
\textit{s.t. } &  P_{\Omega}( \boldsymbol{\mathcal{Y}}) =P_{\Omega}( \boldsymbol{\mathcal{X}}) \nonumber\\ 
&{\mathbf{u}_{r}^{(k)}}^{T} \mathbf{u}_{r}^{(k)}=1,\quad \forall r \nonumber
\end{align}
The penalty $tr(\mathbf{U}_{1}^{T} \mathbf{L}_{i}\mathbf{U}_{1})$is obtained by considering two entities $i_{1}$ and $i_{2}$ with higher similarity (i.e. $a_{i_{1},i_{2}}$ is bigger) should have a closer distance between vector $\mathbf{u}_{i_{1}}$ and $\mathbf{u}_{i_{2}}$ in the matrix $\mathbf{U}_{1}$. When multiple graphs are involved, the formulation can also be easily extended by adding more Laplacian penalties. It is worth noting that the graph definition differs from case to case, and will be clearly explained in the following case study.

\subsection{Efficient Optimization Algorithm}

We propose to use the Block Coordinate Descent (BCD) method to solve our optimization problem. The latent factors and complete tensor to be estimated are divided into $R+1$ blocks:  {$\lambda_{1}, \mathbf{u}_{1}^{(1)}, \mathbf{u}_{1}^{(2)}, ...\mathbf{u}_{1}^{(K)}$}, ... ,  {$\lambda_{R}, \mathbf{u}_{R}^{(1)}, \mathbf{u}_{R}^{(2)}, ...\mathbf{u}_{R}^{(K)}$}, $\mathcal{Y}$. The updating policy is: to update a block of variables which fixing the other blocks, and to one variable which fixing the other variables in each block.

\begin{algorithm}
    \caption{Weakly-dependent on graph LRTC-CP }
  \begin{algorithmic}[1]
    \INPUT incomplete tensor $\mathcal{X}$, $\mathbf{\Omega}$, $\mathbf{L}$, initial rank $R$, maximum iteration $max\_iter$, $\alpha$, $\beta$, $\gamma$, stopping tolerance $tol$.
    \STATE \textbf{Initialization} $\boldsymbol{\mathcal{Y}}_{\Omega}=\boldsymbol{\mathcal{X}}_{\Omega}, \boldsymbol{\mathcal{X}}_{{\Omega}^c}=0$, randomly initialize $\boldsymbol{\lambda}, \{ \mathbf{U}_{k}\}_{k=1,...,K}$, $\boldsymbol{\mathcal{Y}}_{{\Omega}^c}= \{\boldsymbol{\lambda}\times_{1} \mathbf{U}^{(1)}\times_{2} \mathbf{U}^{(2)}...\times_{K} \mathbf{U}^{(K)}\}_{{\Omega}^c}$.
    \FOR{$i = 1$ to $max\_iter$}
      \FOR{$r=1$ to $R$}
        \STATE calculate $\boldsymbol{\mathcal{Y}}_{r}$ by (7).
        \IF{$\lambda_{r} \neq 0$}
          \STATE Update $\mathbf{u}_{r}^{(1)}$ by \cref{eq:u1},  normalize $\mathbf{u}_{r}^{(1)}$.
          \STATE Update $\mathbf{u}_{r}^{(k)}$ by \cref{eq:uk},  normalize $\mathbf{u}_{r}^{(k)}$, $k\neq 1$.
          \STATE Update $\lambda_{r}$ by (14).
          \STATE Update $\boldsymbol{\mathcal{Y}}_{r}$ by \cref{eq:y}.
        \ELSE
          \STATE Do nothing.
        \ENDIF
      \ENDFOR
      \STATE Update $\boldsymbol{\mathcal{Y}}_{\Omega}$ by \cref{eq:y}.
      \STATE Update $R=nonzero(\boldsymbol{\lambda})$.
      \IF{$loss_{i-1}-loss_{i} < tol$}, break; Otherwise, continue.
      \ENDIF
    \ENDFOR
    \OUTPUT $\boldsymbol{\mathcal{Y}}, R, \boldsymbol{\lambda}, \{ \mathbf{U}_{k}\}_{k=1,...,K}$.
  \end{algorithmic}
\end{algorithm}

The Lagrangian function of (6) regarding to the $r$-th block {$\lambda_{r}, \mathbf{u}_{r}^{(1)}, \mathbf{u}_{r}^{(2)}, ...\mathbf{u}_{r}^{(K)}$} is:
 \begin{multline}
 \begin{aligned}
  L_{\lambda_{r}, \mathbf{u}_{r}^{(k)}} & = \frac{1}{2}\|\boldsymbol{\mathcal{Y}}_{r}-\lambda_{r}  \mathbf{u}_{r}^{(1)}\circ \mathbf{u}_{r}^{(2)}\circ\cdots\circ \mathbf{u}_{r}^{(K)}\|^{2}\\
& +\alpha|\mathbf{u}_{r}^{(1)}|+\beta|\lambda_{r}|+ \frac{1}{2}\gamma \cdot tr({\mathbf{u}_{r}^{(1)}}^{T} \mathbf{L} \mathbf{u}_{r}^{(1)})\\ 
 \end{aligned}
 \end{multline}
 where 
 \begin{equation}
    \label{eq:yr} 
\boldsymbol{\mathcal{Y}}_{r}=\boldsymbol{\mathcal{Y}}-\sum_{q\neq r}\lambda_{q} \mathbf{u}_{q}^{(1)}\circ \mathbf{u}_{q}^{(2)}\circ\cdots\circ \mathbf{u}_{q}^{(K)}
 \end{equation} is the residual of the approximation.

\noindent \textbf{Proposition 1}: To  update $\mathbf{u}_{r}^{(1)}$:
\begin{equation}
  \begin{split}
 L_{\mathbf{u}_{r}^{(1)}} &= \frac{1}{2}\|\boldsymbol{\mathcal{Y}}_{r}-\lambda_{r}  \mathbf{u}_{r}^{(1)}\circ \mathbf{u}_{r}^{(2)}\circ\cdots\circ \mathbf{u}_{r}^{(K)}\|^{2} \\ 
 & + \alpha|\mathbf{u}_{r}^{(1)}| + \frac{1}{2}\gamma \cdot tr({\mathbf{u}_{r}^{(1)}}^{T} \mathbf{L} \mathbf{u}_{r}^{(1)}) \\
 & - v \cdot (\mathbf{u}_{r}^{(1)^T}\mathbf{u}_{r}^{(1)}-1)
\end{split}  
\end{equation}

where $v$ is the Lagrangian multiplier. Then partial derivative on $\mathbf{u}_{r}^{(1)}$:
\begin{equation}
\begin{split}
 \frac{\partial L_{\mathbf{u}_{r}^{(1)}}}{{\partial \mathbf{u}_{r}^{(1)}}} = & (\boldsymbol{\mathcal{Y}}_{r}-\lambda_{r}  \mathbf{u}_{r}^{(1)}\circ \mathbf{u}_{r}^{(2)}\circ ...\circ \mathbf{u}_{r}^{(K)}) 
 \\ & \cdot (-\lambda_{r} \mathbf{u}_{r}^{(2)}\circ...\circ \mathbf{u}_{r}^{(K)}) +\alpha\frac{|\mathbf{u}_{r}^{(1)}|}{\partial \mathbf{u}_{r}^{(1)}} \\
 & + \gamma \cdot L \cdot \mathbf{u}_{r}^{(1)} -2v\mathbf{u}_{r}^{(1)}
 \end{split}
 \end{equation}
 By setting the partial derivative as zero,   $\mathbf{u}_{r}^{(1)}$ can be solved in the closed form solution as:
\begin{equation}
\label{eq:u1}
\begin{split}
\mathbf{u}_{r}^{(1)} =& shrink_{\alpha}((\lambda_{r}^{2} \mathbf{I} + \gamma \mathbf{L})^{-1} \\ & \cdot(\lambda_{r}  \boldsymbol{\mathcal{Y}}_{r}\times_{2}\mathbf{u}_{r}^{(2)}...\times_{K}\mathbf{u}_{r}^{(K)}) )
\end{split}
\end{equation}where $shrink_{a}(\cdot)$ is the soft-thresholding operator where $\alpha$ is the regularization parameter.
Then we consider the normalization: $ \mathbf{u}_{r}^{(1)} \leftarrow \frac{\mathbf{u}_{r}^{(1)}}{\|\mathbf{u}_{r}^{(1)}\|_{2}}$.

\noindent \textbf{Proposition 2}: To update  $\mathbf{u}_{k}^{(r)}, k = 2, 3, ..., K $:
\begin{equation}
\label{eq:uk}
\begin{split}
 L_{\mathbf{u}_{r}^{(k)}} =&  \frac{1}{2}\|\boldsymbol{\mathcal{Y}}_{r}-\lambda_{r}  \mathbf{u}_{r}^{(1)}\circ \mathbf{u}_{r}^{(2)}\cdots \mathbf{u}_{r}^{(K)}\|^{2} \\ & -\gamma({\mathbf{u}_{r}^{(k)}}^{T}\mathbf{u}_{r}^{(k)}-1)
 \end{split}
\end{equation}
Similarly we ignore the normalization constraint first by moving normalization step after solving its derivative equation.  By setting the partial derivative as zero, we can have:
 \begin{equation}
  \mathbf{u}_{r}^{(k)}=\frac{\boldsymbol{\mathcal{Y}}_{r}\times_{j}\{\mathbf{u}_{r}^{(j)}\}_{j\neq k}}{\lambda_{r}}
 \end{equation}
\noindent \textbf{Proposition 3}: To update $\lambda_{r}  $:
\begin{equation}
L_{\lambda_{r}} = \frac{1}{2}\|\boldsymbol{\mathcal{Y}}_{r}-\lambda_{r}  \mathbf{u}_{r}^{(1)}\circ \mathbf{u}_{r}^{(2)}\circ\cdots\circ \mathbf{u}_{r}^{(K)}\|^{2}+\beta|\lambda_{r}|
\end{equation}
By setting the derivative on $\lambda_{r}$ as zero, we can have:
\begin{equation}
\begin{split}
 \lambda_{r}=shrink_{\beta}(\boldsymbol{\mathcal{Y}}_{r}\times_{1}\mathbf{u}_{r}^{(1)}\times_{2}\mathbf{u}_{r}^{(2)}\times...\times_{K}\mathbf{u}_{r}^{(K)})
\end{split}
\end{equation}
\noindent \textbf{Proposition 4}: To update $\mathcal{Y}$:
\begin{equation}
\begin{split}
\mathop{\mathrm{min}}_{\boldsymbol{\mathcal{Y}}} & \frac{1}{2}\|\boldsymbol{\mathcal{Y}}-\sum_{r=1}^{R}\lambda_{r}  \mathbf{u}_{r}^{(1)}\circ \mathbf{u}_{r}^{(2)}\circ...\circ \mathbf{u}_{r}^{(K)}\|^{2}\\ 
\textit{s.t. } & P_{\Omega}(\boldsymbol{\mathcal{Y}})=P_{\Omega}(\boldsymbol{\mathcal{X}})
\end{split}
\end{equation}
thus we have:
\begin{equation}
\label{eq:y}
\boldsymbol{\mathcal{Y}}=P_{\Omega}(\boldsymbol{\mathcal{Y}})+P_{\Omega^{C}}(\sum_{r=1}^{R}\lambda_{r}  \mathbf{u}_{r}^{(1)}\circ \mathbf{u}_{r}^{(2)}\circ...\circ \mathbf{u}_{r}^{(K)})
\end{equation}

\section{Experiments}

\noindent In this session, we will implement the proposed model in metro passenger flow data and compare the proposed method with other state-of-art tensor completion methods.

\textbf{Dataset}: The dataset we use is passenger entry and exit data in Hong Kong metro stations.
Each business day is defined as 5:00 AM to 1:00 AM of the next day, and we down-sample it by grouping every 5 minutes, with 247 data points each day. For data split, we use data from 5:00AM, 01-Jan-2017 to 01:00AM, 19-Feb-2017 as observed data (training set), data from 05:00AM, 19-Feb-2017 to 11:10AM, 20-Feb-2017 as validation set and data from 11:10AM, 20-Feb-2017 to 1:00AM, 21-Feb-2017 as missing data (testing set). 
Besides, for data simplification \cite{sun2016understanding}, we randomly choose 15 stations with diverse features.
The 15 stations present quite different passenger flow pattern. So eventually the data we use is: $\boldsymbol{\mathcal{X}}^{L \times P \times T}$ with $L=15$, $P=247$ and $T=51$.

The 15 stations presents quite different inflow passenger flow profile as shown in Figure 2: station 80, located in school area, has strong morning peak (probably for the commuting to school); station 38, locating in business area, has obvious afternoon peak (maybe because of work-off); station 42 in residential area have dual peak for to-and-from home.

\begin{figure}[t]
\centering
\includegraphics[width=0.95\columnwidth]{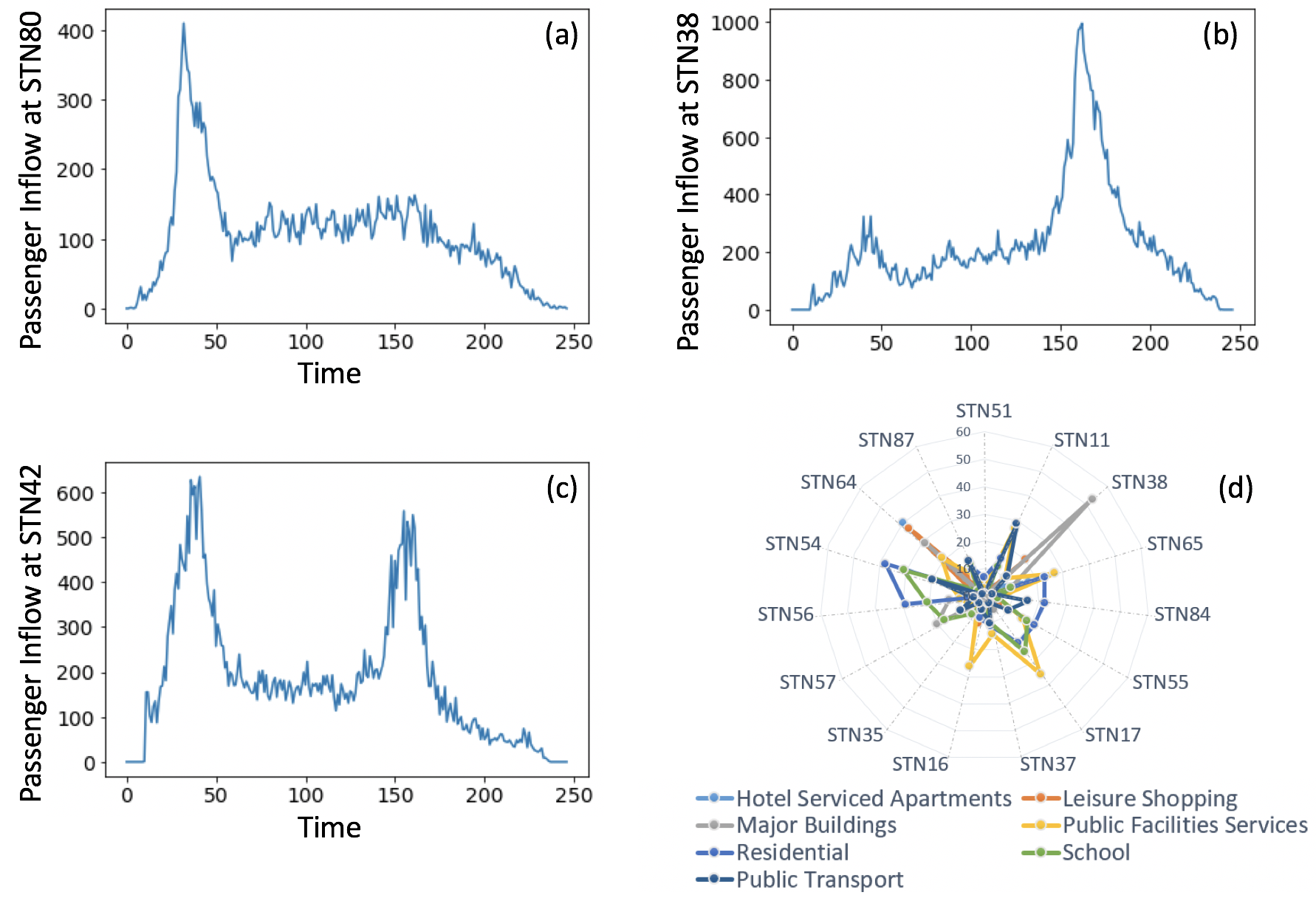} 
\caption{(a),(b),(c) are the passenger flow profile for station 80, 38, 42 at 06-Jan-2017 (Fri) respectively; (d) the POI information of selected stations}
\label{fig2}
\end{figure}

\textbf{Graph}: As mentioned before, the contextual similarity graph and metro network graph play an remarkable role in the passenger flow pattern of each station. If two stations are quite similar in contextual graph or close in metro network graph, then they are more likely to share some passenger flow pattern. For contextual similarity, the Point Of Information (POI) is quite well applied \cite{zhong2017spatiotemporal}, and we denote it as Graph $\mathbf{G}_{POI}$. 
For each station, the information of its surrounding facilities are collected as vector (Hotel Serviced Apartments, Leisure Shopping, Major Buildings, Public Facilities Services, Residential, School, Public Transport), with each dimension value as the total amount of the corresponding facilities. In Figure 2.(d), we can easily observe that, some stations have dominant POI pattern: station 38 is dominant by major office building, thus its passenger flow has afternoon peak pattern.

After obtaining the POI vector of each station, then the POI similarity of station $i$ and $j$ can be calculated as cosine similarity:
\begin{equation}
\{\mathbf{G}_{POI}\}_{i,j}= \frac{\textbf{POI}_{i} \cdot {\textbf{POI}_{j}}}{\|\textbf{POI}_{i}\| \cdot \|\textbf{POI}_{j}\|}
\end{equation}

The metro network graph $\mathbf{G}_{NET}$ can be also defined in several ways: some define it as K-hop, which means how many hops are needed for travelling from station $i$ to station $j$, which is $\{\mathbf{G}_{NET}\}_{i,j}= hop\_amount_{i,j}$; Or some define it as binary graph: when $hop\_amount_{i,j} \leqslant K$ is called 'connected' with label '1', otherwise with label '0', which is shown as following:
\begin{equation}
  \{\mathbf{G}_{NET}\}_{i,j}=\begin{cases}
              1 &  hop\_amount_{i,j} <= K\\
              0 &  hop\_amount_{i,j} > K\\
            \end{cases}
\end{equation}

\textbf{Method for Evaluation}: We compare the proposed model with the following methods for our case.

\begin{table}[t]
\begin{threeparttable}
\caption{Best Parameter Searching}
\label{tab:1}
\setlength\tabcolsep{0pt} 

\begin{tabular*}{\columnwidth}{@{\extracolsep{\fill}} ll cccc}
\toprule
     No. & $\beta$ \tnote{(1)}  & $\alpha$ \tnote{(2)}  & $\gamma$ \tnote{(3)} & $\delta$ \tnote{(4)} & MSE($\times10^6$)\\ 

\midrule
     1&1000 & 50  &70 &80 &3.34\\
     2&1100 & 60  &100 &100 &6.07\\
     3&1200 & 70  &20 &20 &7.98\\
     4&1300 & 30  &0 &80 &7.95\\
     5&1400 & 80  &50 &80 &\textbf{2.68}\\
     6&1500 & 80  &70 &60 &6.53\\
     7&1600 & 30  &10 &0 &4.21\\
     8&1700 & 60  &70 &80 &5.77\\
     9&1800 & 80  &80 &60 &6.94\\
     10&1900 & 30  &100 &20 &5.98\\
     11&2000 & 70  &30 &40 &5.95\\
     12& 1400 & 0 & 0 & 0 & 8.32\\
     13& 1400 & 80 & 0 & 0 & 4.94\\
     14& 1400 & 80 & 50 & 0 & 3.25\\
     15& 1400 & 80 & 50 & 80 & 2.68\\
\bottomrule
\end{tabular*}

\smallskip
\scriptsize
\begin{tablenotes}
\RaggedRight
\item[]  (1) Low-Rank Penalty; (2) Weak Dependency Penalty; (3) $\mathbf{G}_{POI}$ Penalty; (4)  $\mathbf{G}_{NET}$ Penalty
\end{tablenotes}

\end{threeparttable}
\end{table}

\begin{quote}
\begin{itemize}
\item \textbf{Low-Rank Tensor Completion by Riemannian Optimization} (geomCG): which adds nuclear norm on the unfolded tensor along each mode, and performs Riemannian optimization techniques \cite{kressner2014low} 
\item \textbf{High Accuracy Low Rank Tensor Completion} (HaLRTC): which similarly adds nuclear norm on the unfolded tensor along each mode, but solves it by Alternating Direction Method of Multipliers (ADMM) algorithm (Liu et al. 2013).
\item \textbf{Fully Bayesian CP Factorization} (FBCP): which is based on CP decomposition and bayesian graphical model, and assumes that all the decomposed mode matrices are generated from a same Gaussian distribution, whose inverse covariance matrix is generated by another gamma distribution \cite{zhao2015bayesian}. 
\item \textbf{Tensor Rank Estimation based on $L_{1}$-regularized orthogonal CP decomposition} (TREL1\_CP); which achieve the Low-Rank on CP decomposition by introducing $L_{1}$ penalty on weight vector \cite{shi2017tensor}.
\item \textbf{Low-Rank Tensor Decomposition with feature Variance Maximization via CP} (TDVM\_CP): which is based on TREL1-CP and maximizes the feature variance \cite{shi2018feature}.
\item \textbf{Multi-Task for Inflow and Outflow Prediction combining the Lasso regularization term, Ridge regularization term, and Laplacian regularization term} (MTIOP\#LRS): which also introduces POI, traffic capacity, connectivity and weather information as factors into the model, whereas the traffic flow as response  \cite{zhong2017spatiotemporal}.
\end{itemize}
\end{quote}

\subsection{Performance Comparison}

For all methods, we tune the model parameters by grid search according to the performance on the validation dataset and report the performance on the testing dataset.Ther performance metric we consider are Mean Square Error (MSE), Relative Residual (RES) and Mean Absolute Percentage Error (MAPE). Table 2 illustrates the performance comparison of all the mentioned methods. Take the method we proposed in this case study for example, since we introduced two graphs, four parameter are required to be tuned, which is shown as:
$\mathop{\mathrm{min}}_{R, \boldsymbol{\lambda}, {\{\mathbf{u}_{r}^{(k)}\}}} loss  
+ \alpha\|\mathbf{U}_{1}\|_{1}  + \beta\|\boldsymbol{\lambda}\|_{1} 
+ \frac{1}{2}\gamma\cdot tr(\mathbf{U}_{1}^{T} \mathbf{L}_{POI}\mathbf{U}_{1})  + \frac{1}{2}\delta\cdot tr(\mathbf{U}_{1}^{T} \mathbf{L}_{NET}\mathbf{U}_{1})$,
where $\alpha, \beta, \gamma, \delta $ are searched within the grid [0, 100], [1000, 2000], [0, 100] and [0, 100] respectively, with searching step as 10, 100, 10 and 10 respectively. These grids are selected by trial-and-error until we find the satisfactory grid. And the steps are chosen out of the consideration of lowering computational cost and the insensibility of results to the tuning parameters.


According to Table 1, we can make the decision that the parameter combination No. 5 can yield the best completion result; And besides, for each low-rank penalty $\beta$, the best weak dependency penalty $\alpha$, POI graph penalty $\gamma$, and NET graph penalty $\delta$ are almost always nonzero, which means introducing those penalty can truly improve the completion performance. Compared with No.5 with No. 12 to 15, we found that the main contribution to the improvement is  introducing the weakly-correlated penalty, achieving almost 60\% MSE decrease, the POI graph contributing around 30\% MSE decrease, whereas NET graph contributing 10\%. 

\begin{table}[t]
\begin{threeparttable}
\caption{Comparison of Benchmark Methods}
\label{tab:2}
\setlength\tabcolsep{0pt} 

\begin{tabular*}{\columnwidth}{@{\extracolsep{\fill}} ll cccc}
\toprule
     Method& MSE($\times10^6$)& MAPE (\%) \\ 
\midrule
     geomCG & 74.2 &  2881.5 \\
     HaLRTC & 8.41 & 84.12  \\
\addlinespace
     FBCP & 3.98 & 172.81 \\
     TREL1\_CP & 11.19 & 290.85\\
     TDVM\_CP & 8.44 & 180.95 \\
\addlinespace
      MTIOP\#LRS & 6.15 & 116.41 \\
\addlinespace
      WDGTC (our method)& \textbf{2.68} & \textbf{63}\\
\bottomrule
\end{tabular*}
\end{threeparttable}
\end{table}

According to Table 2, We can observe that : (1) geomCG cannot almost yield prediction at all, and the possible reason is that: in our experiment setting all the missing data aggregate in a quite small portion of the tensor, that is: only the data from 11:10AM at last day is missing with missing rate as 1.31\% only, so geomCG is prone to converge on the observation set immediately before starting to estimate missing part; (2) According to Table 1, for the overall MSE, our method can achieve almost 30\% error reduction comparing with the second best baseline method FBCP; (3) Although method MTIOP\#LRS introduces auxiliary information as well, passenger flow does not directly have a statistically functional relationship with all those factors.
\begin{table*}[t]
\centering
\caption{Performance of All Methods on Each Station}\smallskip
\begin{tabular}{l|l|l|l|l|l|l|l}
\hline
(RES) & geomCG &  HaLRTC & FBCP & TREL1\_CP & TDVM\_CP & MTIOP\#LRS & WDGTC\\\hline
STN0  &21.57 &  0.09 & 0.11& 0.13 & 0.13 & 0.11 &\textbf{0.06}\\ 
STN1  & 7.61&  0.14 & 0.19& 0.09 & 0.09 & 0.09 &\textbf{0.08}\\
STN2  & 9.12&  2.71& 0.19 & 3.65 & 3.24 &2.88 &1.13\\
STN3  & 16.93&  0.31& 0.33 & 0.34 & 0.32 & 0.31&\textbf{0.19}\\
STN4  & 22.10&  0.13 & 0.26 & 0.12 & 0.1 & 0.1 & \textbf{0.09}\\
STN5  & 6.50&  \underline{1.52}&\underline{0.76} &\underline{1.25} & \underline{0.68} &\underline{1.12} &\textbf{0.47}\\
STN6  & 8.59 &  0.31 &0.33& 0.39 & 0.28 & 0.26 &\textbf{0.19}\\
STN7  &7.60 &  0.11 &0.12& 0.07 & \textbf{0.06} & 0.07 & 0.1\\
STN8  & 6.78&  \underline{0.87}&\underline{0.57}&\underline{1.17}& \underline{0.56} & \underline{0.54} &\textbf{0.27}\\
STN9  & 20.55 &  0.60 & 0.23 &0.67& 0.64 & 0.47 &\textbf{0.22}\\
STN10 & 10.52 &  1.68& 0.58 & 2.46 & 2.22 & 1.22 &1.03\\
STN11  & 10.29 &  \underline{2.04}& 0.22 & \underline{3.14} & \underline{2.83} & \underline{2.07} &\textbf{0.18}\\
STN12  & 4.31&  0.26& 0.19 & 0.13 & 0.11 & 0.17 &\textbf{0.1}\\
STN13  & 7.07&  3.27& \textbf{0.72} & 5.32 & 3.45 &2.31 &1.79\\
STN14  & 4.07&  0.14 & 0.38 & 0.14 & 0.18 & 0.14 &\textbf{0.11}\\
\hline
\end{tabular}
\label{table1}
\end{table*}

According to Table 3, we can find that: (1) our method can yield the most accurate completion in most of the selected stations (with the best performance in boldface); (2) in some stations, for example, station 5, station 8 and station 11, since its passenger flow pattern is quite unique, so most of other method perform badly (with the bad performance underlined). And the reason is: those method don't complete the missing data obeying the weakly-dependent pattern and graph structure. The completed passenger flow profile by our method, and the second best method (FBCP) are plotted in Figure 3.

\begin{figure}[t]
\centering
\includegraphics[width=0.95\columnwidth]{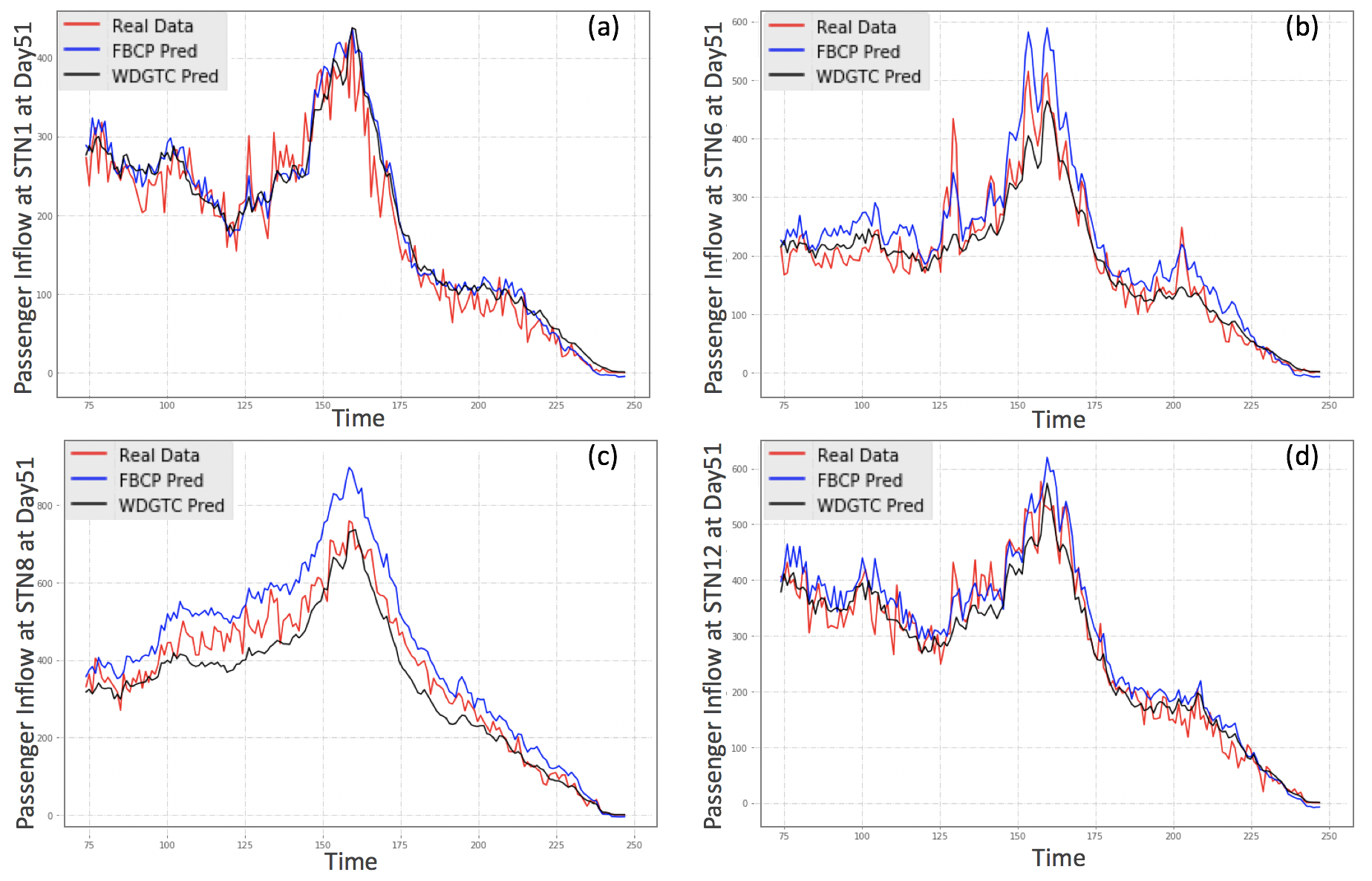} 
\caption{(a), (b), (c) and (d) Completion Result for station 1,6,8 and 12 respectively}
\label{fig3}
\end{figure}

\section{Conclusion and Future Work}
In this paper, we studied a tensor completion problem for a weakly-dependent tensor on graphs. Based on the metro passenger flow tensor, we defined the weakly-dependent pattern and the graph structure behind this. Based on the problem definition, we proposed a novel tensor completion method regularized tensor decomposition by introducing weakly dependent penalty and graph penalty. In addition, we offered a solution for this optimization through Block Coordinate Descent. In the case study based on metro passenger flow data, we defined two graphs which are decisive for each station's pattern. Here, we prove the overall accuracy of the proposed method. For future research, our aim is to formulate the Tucker Decomposition based completion version for the weakly dependent on graph structure. 

\section{Acknowledgments}
This research is supported by Hong Kong MTR Co. with grant numbers RGC GRF 16203917 and 16201718. We also give special thanks to Dr. Qiquan Shi, and Dr. Qibin Zhao for sharing the scripts of their methods.

\bibliographystyle{aaai}
\bibliography{reference.bib}

\begin{thebibliography}{}

\bibitem[\protect\citeauthoryear{Chen, He, and Sun}{2019}]{chen2019bayesian}
Chen, X.; He, Z.; and Sun, L.
\newblock 2019.
\newblock A bayesian tensor decomposition approach for spatiotemporal traffic
  data imputation.
\newblock {\em Transportation research part C: emerging technologies}
  98:73--84.

\bibitem[\protect\citeauthoryear{Chen, Hsu, and
  Liao}{2013}]{chen2013simultaneous}
Chen, Y.-L.; Hsu, C.-T.; and Liao, H.-Y.~M.
\newblock 2013.
\newblock Simultaneous tensor decomposition and completion using factor priors.
\newblock {\em IEEE transactions on pattern analysis and machine intelligence}
  36(3):577--591.

\bibitem[\protect\citeauthoryear{Dunlavy, Kolda, and
  Acar}{2011}]{dunlavy2011temporal}
Dunlavy, D.~M.; Kolda, T.~G.; and Acar, E.
\newblock 2011.
\newblock Temporal link prediction using matrix and tensor factorizations.
\newblock {\em ACM Transactions on Knowledge Discovery from Data (TKDD)}
  5(2):10.

\bibitem[\protect\citeauthoryear{Kolda and Bader}{2009}]{kolda2009tensor}
Kolda, T.~G., and Bader, B.~W.
\newblock 2009.
\newblock Tensor decompositions and applications.
\newblock {\em SIAM review} 51(3):455--500.

\bibitem[\protect\citeauthoryear{Kressner, Steinlechner, and
  Vandereycken}{2014}]{kressner2014low}
Kressner, D.; Steinlechner, M.; and Vandereycken, B.
\newblock 2014.
\newblock Low-rank tensor completion by riemannian optimization.
\newblock {\em BIT Numerical Mathematics} 54(2):447--468.

\bibitem[\protect\citeauthoryear{Lee \bgroup et al\mbox.\egroup
  }{2007}]{lee2007efficient}
Lee, H.; Battle, A.; Raina, R.; and Ng, A.~Y.
\newblock 2007.
\newblock Efficient sparse coding algorithms.
\newblock In {\em Advances in neural information processing systems},
  801--808.

\bibitem[\protect\citeauthoryear{Liu \bgroup et al\mbox.\egroup
  }{2012}]{liu2012tensor}
Liu, J.; Musialski, P.; Wonka, P.; and Ye, J.
\newblock 2012.
\newblock Tensor completion for estimating missing values in visual data.
\newblock {\em IEEE transactions on pattern analysis and machine intelligence}
  35(1):208--220.

\bibitem[\protect\citeauthoryear{M{\o}rup, Hansen, and
  Arnfred}{2008}]{morup2008algorithms}
M{\o}rup, M.; Hansen, L.~K.; and Arnfred, S.~M.
\newblock 2008.
\newblock Algorithms for sparse nonnegative tucker decompositions.
\newblock {\em Neural computation} 20(8):2112--2131.

\bibitem[\protect\citeauthoryear{Shi \bgroup et al\mbox.\egroup
  }{2018}]{shi2018feature}
Shi, Q.; Cheung, Y.-M.; Zhao, Q.; and Lu, H.
\newblock 2018.
\newblock Feature extraction for incomplete data via low-rank tensor
  decomposition with feature regularization.
\newblock {\em IEEE transactions on neural networks and learning systems}
  30(6):1803--1817.

\bibitem[\protect\citeauthoryear{Shi, Lu, and Cheung}{2017}]{shi2017tensor}
Shi, Q.; Lu, H.; and Cheung, Y.-m.
\newblock 2017.
\newblock Tensor rank estimation and completion via cp-based nuclear norm.
\newblock In {\em Proceedings of the 2017 ACM on Conference on Information and
  Knowledge Management},  949--958.
\newblock ACM.

\bibitem[\protect\citeauthoryear{Sun and Axhausen}{2016}]{sun2016understanding}
Sun, L., and Axhausen, K.~W.
\newblock 2016.
\newblock Understanding urban mobility patterns with a probabilistic tensor
  factorization framework.
\newblock {\em Transportation Research Part B: Methodological} 91:511--524.

\bibitem[\protect\citeauthoryear{Wang \bgroup et al\mbox.\egroup
  }{2015}]{wang2015citywide}
Wang, S.; He, L.; Stenneth, L.; Yu, P.~S.; and Li, Z.
\newblock 2015.
\newblock Citywide traffic congestion estimation with social media.
\newblock In {\em Proceedings of the 23rd SIGSPATIAL International Conference
  on Advances in Geographic Information Systems}, ~34.
\newblock ACM.

\bibitem[\protect\citeauthoryear{Yokota, Zhao, and
  Cichocki}{2016}]{yokota2016smooth}
Yokota, T.; Zhao, Q.; and Cichocki, A.
\newblock 2016.
\newblock Smooth parafac decomposition for tensor completion.
\newblock {\em IEEE Transactions on Signal Processing} 64(20):5423--5436.

\bibitem[\protect\citeauthoryear{Yu \bgroup et al\mbox.\egroup
  }{2019}]{yu2019coupled}
Yu, K.; He, L.; Philip, S.~Y.; Zhang, W.; and Liu, Y.
\newblock 2019.
\newblock Coupled tensor decomposition for user clustering in mobile internet
  traffic interaction pattern.
\newblock {\em IEEE Access} 7:18113--18124.

\bibitem[\protect\citeauthoryear{Zhang \bgroup et al\mbox.\egroup
  }{2018}]{zhang2018weakly}
Zhang, C.; Yan, H.; Lee, S.; and Shi, J.
\newblock 2018.
\newblock Weakly correlated profile monitoring based on sparse multi-channel
  functional principal component analysis.
\newblock {\em IISE Transactions} 50(10):878--891.

\bibitem[\protect\citeauthoryear{Zhao, Zhang, and
  Cichocki}{2015}]{zhao2015bayesian}
Zhao, Q.; Zhang, L.; and Cichocki, A.
\newblock 2015.
\newblock Bayesian cp factorization of incomplete tensors with automatic rank
  determination.
\newblock {\em IEEE transactions on pattern analysis and machine intelligence}
  37(9):1751--1763.

\bibitem[\protect\citeauthoryear{Zhong \bgroup et al\mbox.\egroup
  }{2017}]{zhong2017spatiotemporal}
Zhong, R.; Lv, W.; Du, B.; Lei, S.; and Huang, R.
\newblock 2017.
\newblock Spatiotemporal multi-task learning for citywide passenger flow
  prediction.
\newblock In {\em 2017 IEEE SmartWorld, Ubiquitous Intelligence \& Computing,
  Advanced \& Trusted Computed, Scalable Computing \& Communications, Cloud \&
  Big Data Computing, Internet of People and Smart City Innovation
  (SmartWorld/SCALCOM/UIC/ATC/CBDCom/IOP/SCI)},  1--8.
\newblock IEEE.

\end{thebibliography}

\end{document}